\documentclass[nohyperref]{article}

\usepackage{microtype}
\usepackage{subfig}
\usepackage{graphicx}
\usepackage{booktabs} 

\usepackage{hyperref}

\usepackage[accepted]{icml2022}

\usepackage{amsmath}
\usepackage{amssymb}
\usepackage{mathtools}
\usepackage{amsthm}

\usepackage[capitalize,noabbrev]{cleveref}

\theoremstyle{plain}

\theoremstyle{definition}

\theoremstyle{remark}

\usepackage[textsize=tiny]{todonotes}

\icmltitlerunning{Deep Reinforcement Learning for Hierarchically Structured Multi-Agent Warehouse Problems}

\begin{document}

\twocolumn[
\icmltitle{Hierarchically Structured Scheduling and Execution of Tasks in a Multi-Agent Environment}



\icmlsetsymbol{equal}{*}

\begin{icmlauthorlist}
\icmlauthor{Diogo S. Carvalho}{comp,yyy}
\icmlauthor{Biswa Sengupta}{comp}
\end{icmlauthorlist}

\icmlaffiliation{yyy}{INESC-ID \& Instituto Superior Técnico, University of Lisbon}
\icmlaffiliation{comp}{Zebra Technologies, London, UK}

\icmlcorrespondingauthor{Diogo S. Carvalho}{diogo.s.carvalho@tecnico.ulisboa.pt}
\icmlcorrespondingauthor{Biswa Sengupta}{biswa.sengupta@zebra.com}

\icmlkeywords{Machine Learning, ICML}

\vskip 0.3in
]



\printAffiliationsAndNotice{}  

\begin{abstract}
In a warehouse environment, tasks appear dynamically. Consequently, a task management system that matches them with the workforce too early (e.g., weeks in advance) is necessarily sub-optimal. Also, the rapidly increasing size of the action space of such a system consists of a significant problem for traditional schedulers. Reinforcement learning, however, is suited to deal with issues requiring making sequential decisions towards a long-term, often remote, goal.
In this work, we set ourselves on a problem that presents itself with a hierarchical structure: the task-scheduling, by a centralised agent, in a dynamic warehouse multi-agent environment and the execution of one such schedule, by decentralised agents with only partial observability thereof.
We propose to use deep reinforcement learning to solve both the high-level scheduling problem and the low-level multi-agent problem of schedule execution.
Finally, we also conceive the case where centralisation is impossible at test time and workers must learn how to cooperate in executing the tasks in an environment with no schedule and only partial observability.
\end{abstract}

\section{Introduction}
Tasks appear dynamically in a warehouse or retail store environment, and multiple agents must cooperate in executing them. The tasks themselves may consist, for example, of picking items, packing items, cleaning or replenishment, and usually have specific spatial requirements, whether those are the location of the item to be picked or packed, the place where cleaning is required or the shelf in need of replenishment. The dynamics of tasks coming into the environment are random and may be driven by various factors such as online promotions, weather changes, publicity, traffic, international events, and so on. Finally, tasks usually require one or more workers, be they humans or cobots, with a specific skill set to complete them. 
Within the described scenario, we set ourselves on a problem that presents itself with a hierarchical structure with two levels. On the high level, a centralised agent, which we refer to as scheduler, must come up with a schedule, matching decentralised agents, which we refer to as workers, with tasks. On the low level, workers must execute their schedule, respecting its order. The latter challenge composes a multi-agent partially observable Markov game.
\begin{figure}[t]
\centering
\includegraphics[width=0.9\columnwidth]{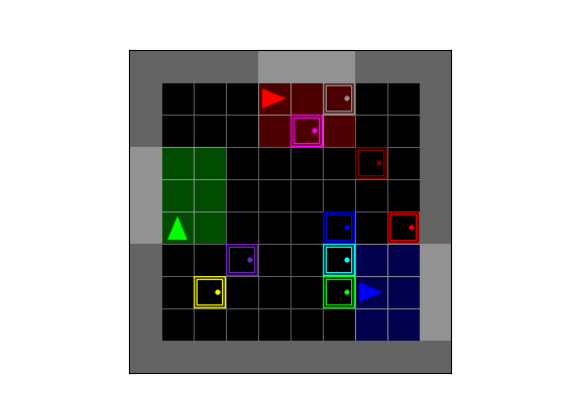}
\caption{$10\times10$ multi-agent grid world environment with 3 agents. Required tasks are represented as doors. Colors identify the agents and the available tasks. Agents have partial observability and can only see a $3$-square around them. Tasks on the backlog do not appear on the environment.}
\label{gym_multigrid}
\end{figure}

Usually, in real-life environments, a human manager or management software that dwells on the high level does not try to capture the task-scheduling problem at the level of granularity we aim at with this work. Instead, workers are usually assigned categories of tasks they should be prepared to execute with days, weeks or months in advance. Consequently, such decisions are sub-optimal, and there often is excess or shortage of workforce to execute tasks, as schedulers fail to adapt to their dynamic environment rapidly. However, continuously changing workers' schedules, at the level of minutes or seconds, may become impractical: firstly, the action space of matching $m$ tasks to $n$ agents that can only take $k$ tasks at a time amounts to $\binom{m}{n \cdot k} \cdot (n\cdot k)!$, which grows as fast as it precludes planning methods, designed to look for optimal solutions; then, more often than not, communication or observability issues are a reality in the environments we consider, especially since we allow for cobots, still prone to failure, to be part of the workforce. Finally, on the low-level environment, we have a decentralised multi-agent problem, which becomes significantly challenging as the state and action spaces grow bigger downstream of increasing environment size, number of agents or tasks.

In this work, motivated by recent successes across the fields of multi-agent, deep and hierarchical reinforcement learning applied to resource management problems that we discuss in the next section; we make the point that hierarchical deep reinforcement learning algorithms are suited to solve the multi-agent warehouse environment problem, both at the upper level of task scheduling and the lower level of task execution. Finally, we also conceive the case where, instead of a two-level hierarchy of a single-agent MDP on the upper level and a multi-agent Markov game on the lower level, we have a one-level decentralised partially observable Markov decision problem, where agents learn to cooperatively execute the tasks in the environment, without centralisation or communication.
The contribution of this work is, therefore, two-fold: (i) a hierarchical multi-agent problem of scheduling and execution of tasks in a simulated warehouse environment; (ii) a deep hierarchical multi-agent reinforcement learning solution to the problem.

The document is outlined as follows. We start by reviewing the relevant literature on related problem settings and hierarchical reinforcement learning; we then provide the necessary background and notation to Markov decision problems, single and multi-agent reinforcement learning and hierarchical reinforcement learning; move to formalising the problem, its hierarchical structure and details of the implementation; describe the approach and present results; conclude and blaze a path for the future.

\section{Related Work}
We split the discussion herein in two, with respect to the problem we address and the framework we fit to it.

\subsection{Resource Management}
One class of problems that have been addressed by planning and reinforcement learning methods and may serve as an umbrella for our issue is the one of resource management. 
Particularly, some problems of resource management can be thought of as machine load balancing \cite{azar1998line}. Planning has been used for commissioning tasks in multi-robot warehouse environments \cite{claes2017decentralised} and tabular reinforcement learning methods were early used for matching professors, classrooms, and classes \cite{ming2010course}. More recently, through deep reinforcement learning, works have explored how to allocate resources such as bandwidth in vehicle-to-vehicle communications \cite{ye2018deep}, power in cloud infrastructures \cite{liu2017hierarchical} and computation jobs in clusters \cite{mao2016resource, chen2017deep, mao2019learning}.

Another pair of problems that intersects with ours is the one of shared autonomous mobility \cite{gueriau2018samod} and particularly fleet management \cite{lin2018efficient}, as tasks and workers in a warehouse environment also have specific spatial attributes, and matching those, the upper level of our problem, has similarities with matching drivers and passengers. Nevertheless, contrarily to ours, work aiming at this problem assume the low-level policies are previously known across drivers, eliminating the added difficulty of a hierarchical reinforcement learning problem that we try to deal with. Different works have approached the high-level problem of dispatching through cascaded learning \cite{fluri2019learning}, both dispatching and relocation of drivers through centralised multi-agent management \cite{holler2019deep}, or just the relocation \cite{lei2020efficient}. Finally, graph neural networks have most recently been used to address the fleet management problem \cite{gammelli2021graph} and the multiple traveling salesman problem \cite{kaempfer2018learning,hu2020reinforcement}.

\subsection{Hierarchical Reinforcement Learning}
Hierarchical reinforcement learning allows learning to happen on different levels of abstraction, structure and temporal extent. The first proposal of a hierarchical framework came through feudal reinforcement learning \cite{NIPS1992_d14220ee}. Managers learn to produce goals to sub-managers, or workers, that learn how to satisfy them. Afterwards, the Option framework \cite{sutton1999between} formalised and provided theoretical guarantees to hierarchies in reinforcement learning. More recently, these ideas have been successfully applied to new algorithms \cite{bacon2017option, nachum2018data}. 

In the specific case of multi-agent systems, after a pioneering work \cite{makar2001hierarchical}, others have explored master-slave architectures \cite{kong2017revisiting}, feudal multi-agent hierarchies \cite{ahilan2019feudal}, temporal abstraction \cite{tang2018hierarchical}, dynamic termination \cite{han2019multi} and skill discovery \cite{yang2019hierarchical}. The field of planning on decentralised partially observable Markov decision processes \cite{oliehoek2016concise} has also seen work leveraging macro-actions \cite{amato2019modeling}.

\section{Background}

We go through the necessary ideas for a hierarchical and multi-agent reinforcement learning setting.

\subsection{Markov Decision Problems}
Reinforcement Learning algorithms propose to solve sequential decision-making problems formally described as Markov Decision Problems \cite{puterman2014markov}. A Markov decision problem is a 5-tuple $(\mathcal{X}, \mathcal{A}, \mathcal{P}, r, \gamma)$, where $\mathcal{X}$ stands for the state space; $\mathcal{A}$ the discrete action space; $\mathcal{P}$ a set of $|\mathcal{A}|$ probability functions, each $P_a: \mathcal{X} \times \mathcal{X} \to [0, 1]$ assigning, through $P_a(x, x')$, the probability that state $x'$ follows from the execution of action $a$ in state $x$; $r: \mathcal{X} \times \mathcal{A} \to \mathbb{R}$ is a possibly stochastic reward function and $\gamma$ is a discount factor in $[0, 1]$.
\subsection{Reinforcement Learning}
In the reinforcement learning setting, the agent does not know the dynamics description $\mathcal{P}$ nor the reward function $r$ of the environment and its goal is to compute the best policy, $\pi^*: \mathcal{X} \times \mathcal{A} \to [0, 1]$, assigning the decision of which action to take in state $x$ in a possibly stochastic manner. Such best policy is defined as the one yielding the biggest amount of summed discounted rewards, in the long run, no matter the state where the agent starts from. To make such notions precise, we define the action value-function of a given policy $\pi$ as $Q^\pi: \mathcal{X} \times \mathcal{A} \to \mathbb{R}$ such that \begin{equation}Q^\pi(x, a) = \mathbb{E}[\sum_{t = 0}^\infty \gamma^t r(x_t, a_t) \mid x_0 = x, a_0 = a],\end{equation} where $x_{t+1}\sim P_{a_t}(x_t, \cdot)$ and $a_t\sim \pi(x_t, \cdot)$. To solve a Markov decision problem, one possible partition of the set of reinforcement learning algorithms that do not explicitly try to learn a model of the world itself, called model-free, breaks them into two: value-based methods and policy-based methods.

On the one hand, value-based methods approximate $\pi^*$ implicitly while approximating its action value-function $Q^*$, known to verify the fixed point equation 
\begin{equation}
    Q^*(x, a) = \mathbb{E}[r(x, a) + \gamma \max_{a' \in \mathcal{A}} Q^*(x', a')].
\end{equation}
The most well-known method to perform such approximation is $Q$-learning \cite{watkins1992q}, a tabular method that founds more sophisticated algorithms such as deep $Q$-networks \cite{mnih2015human} and upgrades over it \cite{hessel2018rainbow, van2016deep, wang2016dueling, fortunato2017noisy, schaul2015prioritized}. The incorporation of deep neural network architectures to allow for information generalisation along the state space was crucial to avoid the curse of dimensionality \cite{burton2010coping}. 
%

On the other hand, policy-based methods are based on direct improvements over the policy. They do require an explicit representation of a policy $\pi$, known as the actor, commonly added of an estimate of its current value, known as the critic. For that reason, most policy-based algorithms are known as actor-critic algorithms \cite{barto1983neuronlike, konda2000actor} and rely on the policy-gradient theorem \cite{sutton2000policy} for the stochastic gradient updates. Letting 
\begin{equation}
    J(\pi) = \mathbb{E}_{x_0\sim \mu}[Q^{\pi}(x, a)],
\end{equation}
where $\mu$ denotes the initial state distribution, the theorem states that
\begin{equation}
    \nabla{J(\pi)} = \mathbb{E}[\nabla \log \pi(x, a) (Q^{\pi}(x,a) - b(x)],
\end{equation}
where $b$ is a baseline that does not bias the gradient estimation but can reduce or increase the variance.
PPO is one such actor-critic methods, where the baseline used is the state value function $b(x) = \mathbb{E}[Q(x, a)]$, added of the optimisation of a surrogate objective loss. Other well-known actor-critic methods are A3C \cite{mnih2016asynchronous}, its synchronous version A2C, DDPG \cite{lillicrap2015continuous}, IMPALA \cite{espeholt2018impala} and SAC \cite{haarnoja2018soft}.

%

While theoretically known to perform well in tabular settings and empirically known to perform well on reasonably tricky problems, the methods described above may struggle in problems with sizeable discrete action spaces. In particular, learning in a fully centralised multi-agent problem through vanilla deep reinforcement learning algorithms, as described above, soon becomes intractable due to the exponential growth of the action-space as the number of agents grows.

\subsection{Multi-Agent Markov Decision Problems}
A decentralised partially observable Markov decision problem (Dec-POMDP) describes a fully cooperative multi-agent environment. More formally, it consists of a tuple $([n], \mathcal{X}, \mathcal{U}, \mathcal{P}, r, \gamma, \mathcal{Z}, O)$, $[n]=(1, 2, \ldots, n)$ representing the indexes of n agents. Identically to the way we formalised an MDP, in a dec-POMDP $\mathcal{X}$ consists of the set of states; $\mathcal{U} = \times_{i \in [n]} \mathcal{A}_i$, where $\mathcal{A}_i$ is the set of actions available for agent $i$; $\mathcal{P}$ is a set of transition probability functions, mapping a triple $(x, u, x') \in \mathcal{X}\times \mathcal{U} \times \mathcal{X}$ to the probability that state $x'$ follows from executing action $u$ in state $x$; $r$ is the immediate reward function, shared across agents, mapping pairs $(x, u)$ to possibly stochastic rewards in $\mathbb{R}$. $\mathcal{Z} = \times_{i \in [n]} \mathcal{O}_i$ is the set of joint observations, where $\mathcal{O}_i$ is the set of observations for agent $i$, and $O$ maps triples $(z, x, u) \in \mathcal{Z} \times \mathcal{X} \times \mathcal{U}$ to the probability that the joint observation $z$ is emitted after the execution of action $u$ in state $x$. 

A partially observable Markov game \cite{hansen2004dynamic} is a more general setting than the one of dec-POMDPs, where rewards are not necessarily shared across agents. Consequently, a partially observable Markov game allows for modelling scenarios that are not fully cooperative, such as competitive or neutral ones. To formalize the relation between the two settings, a partially observable Markov game is a tuple $([n], \mathcal{X}, \mathcal{U}, \mathcal{P}, \mathcal{R}, \gamma, \mathcal{Z}, O)$, where $\mathcal{R}$ is a set of $n$ reward functions $r_i: \mathcal{X} \times \mathcal{U} \to \mathbb{R}$, one for each agent. 

\subsection{Multi-Agent Reinforcement Learning}
Trying to address the multi-agent setting, methods have been developed that rely on either a fully-decentralised approach or through centralised training with decentralised execution, an approach firstly considered in planning \cite{kraemer2016multi} but that has become central in the reinforcement learning literature in the most recent years \cite{NIPS2016_c7635bfd}. At the halfway point between the two approaches (i.e., fully-decentralized and fully-centralized), we can also consider a multi-agent setting where multiple homogeneous agents can share and train a single policy, which can then be decentralised at test time. We refer to the last approach as parameter sharing \cite{gupta2017cooperative}.

The fully-decentralised approach has surprisingly had a few success stories. However, it is still tough for agents to learn independently due to the non-stationarity \cite{papoudakis2019dealing} of the environment caused by other agents' initial behaviour. Any single-agent reinforcement learning method can be implemented this way, including recent deep-RL methods, be them value-based, such as DQN \cite{mnih2015human}, or policy-based, such as PPO \cite{schulman2017proximal}, by simply letting each agent learn as if it were in a single-agent setting.

The centralised training with a decentralised execution approach is, for scenarios where it is applicable, very promising. Most methods consist of a decentralised actor (policy) with a centralised critic (value-function) on learning time. On test time, only the actor is required, thus rendering the methods decentralised. Under the described umbrella, we can fit again many deep-RL methods, either value-based, such as VDN \cite{sunehag2017value}, QMIX \cite{rashid2018qmix}, QTRAN \cite{son2019qtran} and REFIL \cite{iqbal2021randomized}; or policy-based, such as MADDPG \cite{NIPS2017_68a97503}, COMA \cite{foerster2018counterfactual}, MAAC \cite{iqbal2019actor} and SEAC \cite{NEURIPS2020_7967cc8e}. As the primer benchmark for most of the multi-agent reinforcement learning algorithms, StartCraft MultiAgent Challenge has been the most common choice \cite{samvelyan19smac}.

Finally, also parameter-sharing can be used to learn a single policy with any deep-RL algorithm for homogeneous multi-agent settings. The approach has been shown to mitigate the adverse effects of non-stationarities that arise when trying to learn fully decentralised independent policies \cite{terry2020revisiting,pmlr-v139-christianos21a}.

\subsection{Hierarchical Reinforcement Learning}
Hierarchical reinforcement learning, in its turn, has long been promising to contribute to dealing with complex problems, particularly ones with large action spaces. Hierarchically, a single agent may learn to solve smaller, decomposed problems that would be intractable if dealt with on their whole. Options also allow for temporal-extension of actions and their abstraction, leading to a structure with a similar notion as the one of an MDP, called a semi-MDP. 

Among the frameworks that have formalised some interpretation of hierarchy within reinforcement learning, such as the one of feudal learning \cite{NIPS1992_d14220ee}, hierarchical abstract machines \cite{parr1998reinforcement} and MAXQ decomposition \cite{dietterich2000hierarchical}, the most popular one is the options formulation \cite{sutton1999between}. Options extend the traditional definition of actions and Markov decision problems in the sense that, on top of the latter, an option also consists of a termination condition $\zeta: \mathcal{X} \to [0, 1]$ and an initiation set $\mathcal{I} \subseteq \mathcal{X} $, determining respectively when the macro-action ends and when it starts.

Under the lights of hierarchical and multi-agent reinforcement learning formulations, in the following section, we detail how we formalise the warehouse problem in hands.

\section{Problem Setting}
%
In our setting, the warehouse, location and specifications of tasks and agents are modelled as a multi-agent grid world. At each step, a cell of the grid world can be non-occupied, occupied with an agent or occupied with a task. Agents and tasks both have a set of attributes. In the case of agents, they have a colour identifier, a direction, and a schedule composed of their assigned tasks. In the case of tasks, they also have a colour identifier. Tasks appear randomly in the environment on each time step according to independent and identically distributed Bernoulli($p$) random variables. The environment can only take up to $m$ visible tasks, with $m$ denoting the maximum size of the queue. If the queue is full and a new task arrives, it is added to a backlog with no limit number of tasks. When the queue is full, and a task is completed, another from the backlog is added to its place.
%
\subsection{High-level}
On the high-level, an agent fully observes the state of the environment, including the location of tasks and their attributes, the location of agents, their attributes and their current schedules and the number of tasks in the backlog, and produces an action corresponding to the schedule each low-level agent must perform. Each high-level action is, therefore, a tuple in $m^{k \cdot n}$, where $m$ is the size of the queue, $k$ is the size of each low-level agent's schedule, and $n$ is the number of low-level agents. 

The size of the action space, as described above, grows exponentially both on the size of the schedule $k$ and the number of low-level agents $n$. To keep the action space linear in such quantities, the high-level agent recursively chooses one of $m$ tasks to add to a low-level agent's schedule until either its buffer is full or the null action is chosen. Therefore, the high-level agent must output up to $m \cdot k$ tasks to output each schedule.

The high-level agent is rewarded the negative average job slowdown, which can be computed as
\begin{equation}
C = -\sum_{j \in (\mathcal{M} \cup \mathcal{L})} {(D_j)}^{-1},
\end{equation}
where $\mathcal{M}$ is the queue, $\mathcal{L}$ is the backlog, and $D_j$ is the estimated time for actually executing job $j$.
Also, the high-level agent to learn not to add tasks to the schedule that are either already scheduled or not present in the queue also receives a negative unit reward after doing so.

Finally, the high-level agent is called to compute and communicate the new schedule to low-level agents when they have reached a maximum number of actions on the environment.
\subsection{Low-level}
The low level is a decentralised, partially observable multi-agent problem. Since the reward is not shared across agents, the problem is, in particular, a Markov game. Consequently, the low-level does not present itself with a cooperative nature, and there is no incentive for cooperation. However, competition, in a more intuitive sense, is also not particularly incentivised in our setting.

Each agent can see a $v$-square of cells around it, its location and direction and the location of tasks on its schedule and act without communicating with the other agents. At each time step, it is rewarded for the negative ratio of its progress, as the number of tasks of the schedule it still has to perform is divided by the total number of tasks it has been assigned, $k$.

\subsection{Dec-POMDP}
In the dec-POMDP setting, we assume scheduling, and consequently, a hierarchy, is not necessary or impossible. Therefore, workers must cooperate in executing tasks in their environment using only their partial observability and learning through their shared reward. We do allow, however, that during training, centralisation is possible. Notice that, while similar to the low-level of the hierarchical problem above, the decentralised POMDP has two added difficulties: agents do not have the location of tasks they should perform, only finding them once the tasks are inside their observation; agents share the reward, which makes the problem less stationary as rewards depend on other agents' actions, on top of their own.

\subsection{Implementation Details}
For all the algorithmic and learning implementation, we use the RLlib library \cite{liang2018rllib}. To implement the warehouse environment, we build on an existing repository \cite{gym_minigrid,gym_multigrid} allowing for customisation of Gym \cite{1606.01540} compatible multi-agent grid world environments. Figure~\ref{gym_multigrid} showcases a renderization of the environment.

\subsubsection*{Grid}
The custom grid world is an $h \times w$ grid, with $h$ denoting height and $w$ denoting width. Each cell is then composed of a tuple encoding its content. The outer cells are walls that the agent cannot stand on. The inner cells have either wall, an agent or a task. 

\subsubsection*{Tasks}
Tasks are represented as doors and have a certain colour, which identifies them. The task is complete, and the door disappears once an agent opens the corresponding door. To open a door, an agent must only navigate to the door and toggle it. The door then disappears from the environment. The environment may only contain, at a particular time step, $m$ tasks at most. Any additional tasks are in a backlog. The number of tasks in the backlog is visible to the high-level agent, but its content is not. 

\subsubsection*{Agents}
The cells containing one of the $n$ low-level agents, represented as isosceles triangles accounting for the direction they face, contain a tuple with its color, its direction and the tasks it is scheduled to perform. 

\subsubsection*{Episodes}
At the beginning of an episode, there are $m_0$ tasks in the environment. After each low-level step, a new task arrives with probability $p$.


\subsubsection*{Low-Level}
Each low-level agent observes a $v$-sized grid around it, its current direction and the position of the the tasks it is scheduled to perform and must output an action in $\mathcal{A} = \{\textrm{do nothing}, \textrm{move front}, \textrm{turn right}, \textrm{turn left}, \textrm{toggle}\}$.

\subsubsection*{High-Level}
The high-level agent must provide an action, a complete schedule, consisting of $n$ sub-schedules of $k$ tasks each, one for each low-level agent. Such a schedule is recursively computed to keep the action space linear in the number of tasks $m$.

\subsubsection*{Dec-POMDP}
In the decentralised POMDP setting, the reward of the environment is shared across agents, but there is no two-level hierarchy---no schedule. Agents must learn how to coordinate and perform tasks without communication at test time. Agents in the dec-POMDP setting are modelled the same way as low-level agents in the hierarchical problem defined above: each agent observes a $v$-square around it and must execute an action in $\{\textrm{do nothing}, \textrm{move front}, \textrm{turn right}, \textrm{turn left}, \textrm{toggle}\}$. The reward is shared across all agents and equals the negative job slowdown--- the same as the high-level reward in the hierarchical problem.

\section{Evaluation}

\begin{figure*}[ht]
\centering
\subfloat[Fixed buffer size, k=2.]{\label{figur:1}\includegraphics[width=0.5\textwidth]{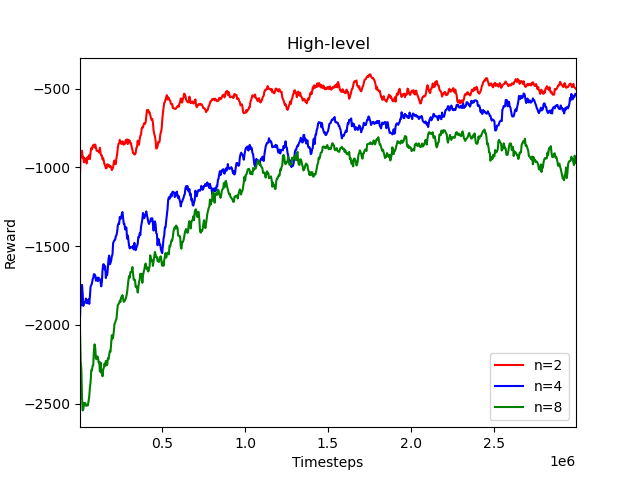}}
\subfloat[Fixed number of agents, n=4.]{\label{figur:2}\includegraphics[width=0.5\textwidth]{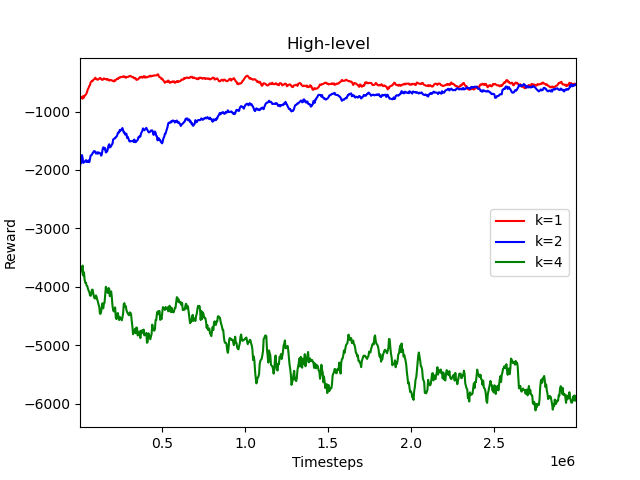}}
\caption{High-level policy undiscounted episodic reward during training after $1\cdot10^6$ time-steps of low-level policy pre-training. Episodic rewards appear on the vertical axis, time-steps on the horizontal axis, regardless of happening on the higher or lower levels. In Figure~\ref{figur:1}, each agent's buffer size $k$ is fixed and $n$ varies. In Figure~\ref{figur:2}, the number of agents $n$ is fixed and $k$ varies.}
\label{high_level_33}
\end{figure*}
Along the section, we are set on a $10\times10$ grid where low-level agents observe a $3$-square around them. We set the probability that a new task appears at a certain time-step as $p=(\frac{n \cdot k}{4} + 1)\%$. Finally, the maximum number of tasks visible on the environment is set as $m = m \cdot k + 1$. The initial number of tasks on the environment, at the beginning of an episode, is randomly and uniformly chosen as $m_0\sim U(\{0, \ldots, m\})$. An episode ends once $4 \cdot h \cdot w$ low-level steps are made. A low-level episode ends once $h \cdot w$ episodes are made. 

\subsection{High-Level}\label{sec:eval-high-level}
We train the high-level agent with the PPO algorithm. The high-level policy network is composed of a convolutional neural network with a ReLU activation function. To leverage the power of the convolutional layers on RLlib, we zero-pad the grid world into a $h \times w \times f(k)$ tensor, $f: \mathbb{N} \to \mathbb{N}$ and each of the $h \cdot w$ $f(k)$-tuples are encoded as described before, with the attributes of agents or tasks.

Training both hierarchical levels at the same time may produce significant non-stationarities that may harm learning. Therefore before starting the high-level training, we train the low-level agents to perform randomly chosen schedules on a static environment ($p = 0$). After training the low-level agent for $10^6$ time steps, we start training the high-level agent. Notice here that we allow the low-level policy to keep training. Training for the high level is performed using the PPO algorithm. In Section~\ref{sec:additional} we present results that validate our choice.

To examine how the high-level scheduler performs and scales with the number of agents, $n$, and the number of tasks each agent can be assigned, $k$, we show learning plots for $n\in\{2, 4, 8\}$ when $k=2$ and $k\in\{1, 2, 4\}$ when $n=4$ on Figure~\ref{high_level_33}, the main set of results in the paper. Notice that, as $n$ or $k$ increases, the frequency  $p$ at which new tasks appear on the environment increases as per our construction, and as described above, rendering the environment more challenging when the workforce is also more capable. Figure~\ref{figur:1} shows that, while the number of agents increases, the method can still learn a scheduler. Figure~\ref{figur:2} shows that while each agent's maximum number of tasks assigned increases, the method may fail to learn a scheduler, particularly for $k=4$. However, we hypothesise that the results are due to the difficulty of the environment itself---tasks appearing too frequently for $n=4$ agents.

\subsection{Low-Level}
Low-level agents are trained with the PPO algorithm with shared experience unless otherwise noted. The low-level policy networks are fully connected networks with a Tanh activation function. Learning hyperparameters are also default for the PPO algorithm. 

\begin{figure}[h]
\centering
\includegraphics[width=0.9\columnwidth]{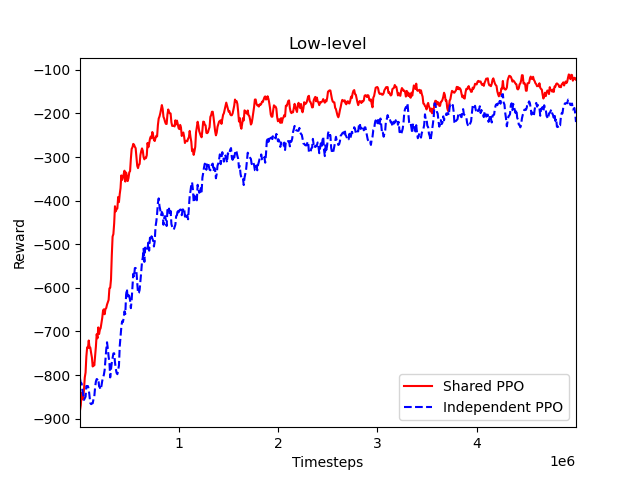}
\caption{Low-level policy training; $n=3, k=3$ and $p=0.05$. Comparison between fully independent and parameter sharing actor-critic algorithm (PPO). The vertical axis shows low-level episodic rewards, the horizontal axis shows low-level time-steps on the environment.}
\label{low_level_33}
\end{figure}

We test the performance of the multi-agent low-level problem using two PPO settings. In one of them, each agent learns its own policy, mapping observations to actions, independently of the others. In the second setting, agents can share experience and the policy, which we call shared-experience PPO. Figure~\ref{low_level_33} shows the results for $n=3$ and $k=3$, with $p=0.05$.
We observe that agents learning with shared experience can learn quicker than independently. 

\subsection{Dec-POMDP}
\begin{figure}[h]
\centering
\includegraphics[width=0.9\columnwidth]{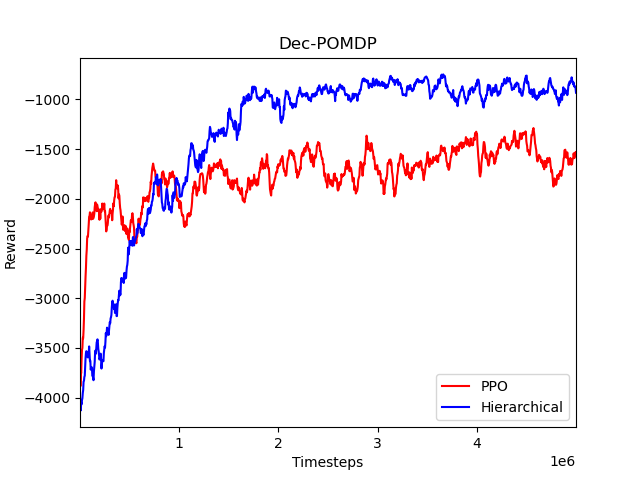}
\caption{Decentralised POMDP training; $n=3$, $k=3$ and $p=0.05$. Episodic rewards are shown as a function of time-steps. The red line shows the environment reward as agents learn a parameter-sharing actor-critic policy (PPO). The blue line shows the high-level reward of the hierarchically structured problem.}
\label{dec-pomdp}
\end{figure}

We test learning through decentralised training with a decentralised execution approach, notably the PPO algorithm with a shared experience. We train the agents on an environment with, again $n=3$ agents and $k=3$ tasks, with $p=0.05$. Figure~\ref{dec-pomdp} shows the results. We added the results from the hierarchical, centralised setting for intuition purposes and should not be directly compared, as the dec-POMDP setting may be considered harder: agents act in a decentralised manner and are unaware of the location of tasks unless they appear in their partial observation. From Figure~\ref{dec-pomdp}, we observe that the agents learn very quickly a decent performance. However, their performance does not improve as well as in the hierarchical problem as training continues.

\subsection{Additional Experiments}\label{sec:additional}
\begin{figure*}[h]
\centering
\subfloat[Low-level pre-training.]{\label{mixed}\includegraphics[width=0.33\textwidth]{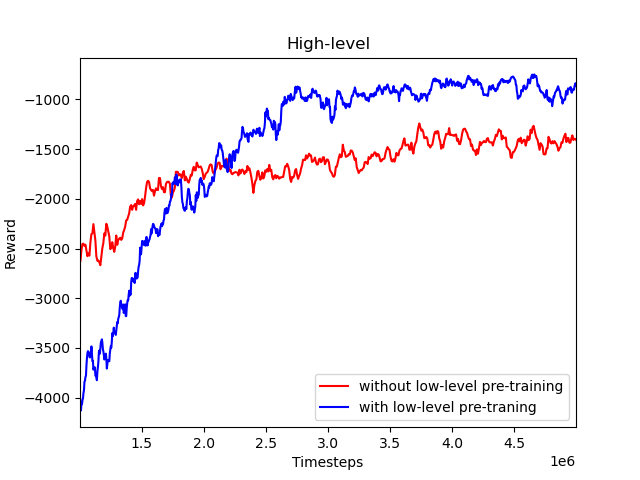}}
\subfloat[Baseline;]{\label{high_level_33_add}\includegraphics[width=0.33\textwidth]{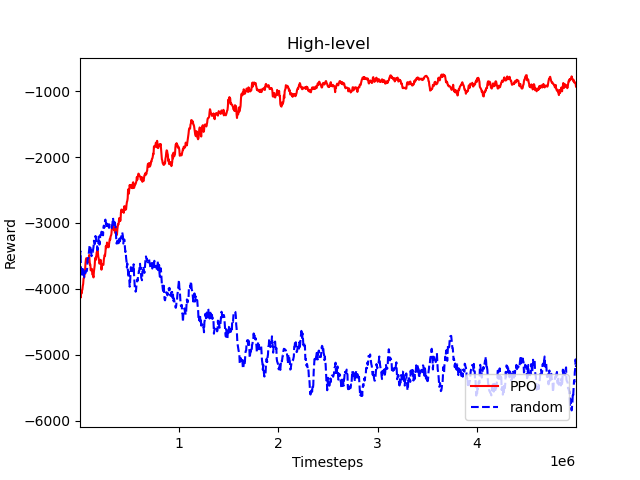}}
\subfloat[Large buffer size. $k=4$; $p$ decreases;]{\label{large_buffer}\includegraphics[width=0.33\textwidth]{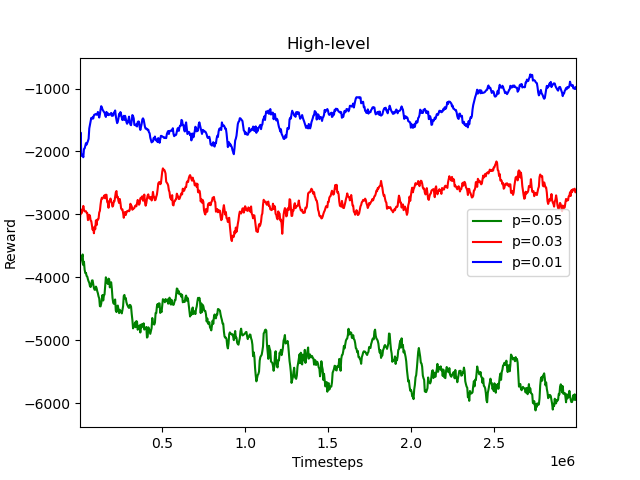}}
\caption{Additional experimental results. Vertical axis shows episode reward; horizontal axis time-steps taken.} 
\label{supp}
\end{figure*}

We hereby include additional experimental results to the proposed hierarchical model for scheduling and executing tasks in a multi-agent warehouse environment described in the main document. The experiments add three insights: effects of low-level policy pre-training; performance of the high-level scheduler against a random baseline; performance of the high-level scheduler for a high buffer-size but as the difficulty of the environment itself decreases.
Recall that the buffer size $k$ is the number of tasks the high-level scheduler must assign to each agent and is consequently the number of tasks the low-level agents must perform until new tasks are assigned to it. Results are described in Figure~\ref{supp}; the settings' details and analysis appear below.

\subsubsection*{Low-level Pre-Training}
To confirm the benefits of the early training of the low-level policy, as described in Section~\ref{sec:eval-high-level}, we compare the results of training both policy levels simultaneously against pre-training the low-level. We show the training plot after the first $1\cdot10^6$ time steps in Figure~\ref{mixed}. As expected, due to the non-stationarity induced by training the two levels simultaneously, learning is worse without pre-training the low-level policy with a random scheduler in a stationary environment.

\subsubsection*{Baseline}
We are set on an environment of $n=3$ agents, each with a buffer-size of $k=3$ and a probability of new tasks coming in to the environment at a new time-step of $p=0.05$. We compare the performance of the method we propose against a random scheduler, having pre-trained the low-level agents in both cases. Figure~\ref{high_level_33_add} shows the results. Clearly, the performance of our method surpasses the one of a random scheduler.

\subsubsection*{Large Buffer Size}
Here, we examine if, when the buffer-size is large, $k=4$, and the dynamics of the environment decrease ($p$ decreases), the performance of the model increases. The environment is, as in the main document, of $n=4$ agents and $p$ varies in $\{0.01, 0.03, 0.05\}$. Figure~\ref{large_buffer} shows the results.
We observe that, as expected, as the difficulty of the environment decreases ($p$ decreases), the method is able to provide a better policy.

\section{Conclusion}
Through this work, we contributed to a hierarchically structured problem of warehouse distribution and execution of tasks and a hierarchical deep reinforcement learning solution to the problem. On a high level, the problem resembles the one of load balancing or fleet management. On the low level, it is a homogeneous multi-agent problem. For both levels, separately, we have tested different approaches. Finally, we have also conceived the case where, in the case of choice or necessity, centralised scheduling is not possible and a fleet of agents with partial observability learns how to execute tasks in the environment cooperatively. The topic and contribution is relevant to the scientific communities of reinforcement learning and operations research and has possible real-life applications in industry.

While the hierarchical multi-agent problem of scheduling and executing tasks is complex enough to be challenging, a warehouse environment is even more. In the future, besides considering larger warehouses, it would be interesting to have heterogeneous fleets of low-level agents and tasks with different skill sets and requirements, respectively. Another difficulty would be the requirement of multiple agents for executing a single task, which would require a considerable level of coordination between agents. It would also be interesting to consider that different tasks may be associated with varying levels of priority. Finally, besides the dynamics of the tasks coming into the environment, the workforce dynamics would also be worth considering: the number of agents may vary with time due to different reasons. To overcome the problems the added difficulties would imply, learning-wise, it may be valuable to add additional levels to the hierarchy, learn a latent embedded representation of actions in a continuous space or exploit the structuring capabilities of graph neural networks.

\bibliography{biblio}
\bibliographystyle{icml2022}


\end{document}